\pdfoutput=1

\documentclass[11pt]{article}

\usepackage{acl}

\usepackage{acl}
\usepackage{times}
\usepackage{latexsym}
\usepackage{stfloats}
\usepackage[T1]{fontenc}

\usepackage[utf8]{inputenc}

\usepackage{microtype}

\usepackage{verbatim}
\usepackage{tikz}
\definecolor{royblue} {HTML}{002484}
\usepackage{subfig}
\usepackage{amssymb}
\usepackage{amsmath}
\newcommand*\rot{\rotatebox{90}}
\usepackage{rotating}
\usepackage{multirow}

\usepackage{arydshln}
\DeclareMathOperator{\LN}{LN}

\usepackage[disable]{todonotes}
\newcounter{notecounter}
\newcommand{\jl}[1]{%
    \stepcounter{notecounter}%
    \todo[inline,color=magenta!40!gray!20]{\scriptsize JL \fbox{\texttt{\footnotesize \arabic{notecounter}}} \footnotesize #1}}

\newcommand{\pjr}[1]{%
    \stepcounter{notecounter}%
    \todo[inline,color=yellow!80!gray!20]{\scriptsize PJR \fbox{\texttt{\footnotesize \arabic{notecounter}}} \footnotesize #1}}

\title{Probing the Role of Positional Information in Vision-Language Models}

\author{Philipp J. Rösch$^1$ \and Jindřich Libovický$^2$ \\
$^1$Institute for Distributed Intelligent Systems, Bundeswehr University Munich, Germany \\
$^2$Faculty of Mathematics and Physics, Charles University, Czech Republic \\
\texttt{philipp.roesch@unibw.de} \quad \texttt{libovicky@ufal.mff.cuni.cz}
}

\begin{document}
\maketitle
\begin{abstract}
    %
In most
Vision-Language models (VL), the understanding of the image structure is enabled by injecting the position information (PI) about objects in the image.
%
%
In our case study of LXMERT, a state-of-the-art VL model, we probe the use of the PI in the representation and study its effect on Visual Question Answering.
We show that the model is not capable of leveraging the PI for the image-text matching task on a challenge set where only position differs.
Yet, our experiments with probing confirm that the PI is indeed present in the representation.
We introduce two strategies to tackle this: (\textit{i}) Positional Information Pre-training and (\textit{ii}) Contrastive Learning on PI using Cross-Modality Matching.
Doing so, the model can correctly classify if images with detailed PI statements match.
Additionally to the 2D information from bounding boxes, we introduce the object's depth as new feature for a better object localization in the space.
Even though we were able to improve the model properties as defined by our probes, it only has a negligible effect on the downstream performance.
Our results thus highlight an important issue of multimodal modeling: the mere presence of information detectable by a probing classifier is not a guarantee that the information is available in a cross-modal setup.
%
\end{abstract}

\jl{Alternative titles: ``\emph{Probing the Role of Positional Information in Language-Vision Models}'', ``\emph{Do we need Positional Information in Language-Vision Models?}'' The word ``importance'' a priori suggest it \emph{is} important, but I guess we are going to say it is more complicated. \pjr{Now I favor version 1.}}

\section{Introduction}

Pre-trained Vision-Language models
\citep{Tan2019LXMERT,Lu2019VILBERT,Yu2020ERNIEVIL,chen2020UNITER} reached
strong performance in many multimodal tasks such as Visual Question Answering \citep{Agrawal2015VQA,Hudson2019GQA,Bigham2010VIZWIZ} or Visual
Inference \citep{Johnson2016CLEVR,Suhr2019NLVR2}.
%
%
All these models use the Transformer architecture \citep{Vaswani2017} and
make use of several pre-training strategies like
\textit{Masked Cross-Modality Language Modeling} (MM) and \textit{Cross-Modality Matching} (CMM) similar to masked language modeling  and next sentence prediction \citep{Devlin2018} in NLP\@.

%
%

Because the attention mechanism treats its inputs as unordered sets, Transformer-based NLP models need to use position encodings to represent the mutual position of the tokens, so that the models can grasp the sentence structure.
The mutual position of objects is equally important for understanding the structure of an image.
VL models differ in how they represent objects in the image, typically represented as sets of object features and PI. 
Therefore, object detectors are 
used to obtain bounding box information for all objects. In many models, the upper left and lower right corners of the object's bounding box are used as 2D information to create a learnable positional encoding.
%
In addition to the spatial but flat 2D values, we determine the depth of the objects in the image and make it available as an additional  feature.
Until now, VL models recognize the objects on a flat map but not in the real three-dimensional context.

We found that the current LXMERT model is capable of forwarding PI
through
the model but is not capable to use it to solve image-text matching tasks where positional keywords are replaced by their counterparts.
Introducing two new pre-training strategies, we target these unimodal and multimodal evaluation schemes and improve probing results.
Yet, we did not get any perfomance increase on the downstream tasks.
This is most likely
due
to the small fraction of position-related text in the pre-training corpus and suboptimal results of the object detector.
Regarding PI type, it seems to be sufficient to input object center values which is far less than most VL model input today.

\section{Positional Information in VL Models}

In NLP, the importance of word order is given great attention \citep{Ke2020rethinking, Wang2020what}. Different methods exist, including analytical position encodings \citep{Vaswani2017}, learnable additive embeddings \citep{Devlin2018} or the relative the attention query \citep{shaw-etal-2018-self}. \pjr{"relation-aware attention query"  instead of "relative the attention query"? Is that correct English?}
There is no equivalent research that would specifically approach PI in VL
models. However, the position of the objects is considered in almost all common Transformer-based approaches.

In LXMERT \citep{Tan2019LXMERT} the upper left and lower right corners of the object
are used to encode its position.
The same is true for M4C \citep{Hu2019M4C}.
Other models also use the relative area fraction of the objects as an additional feature. Although the network should be able to determine this feature, it is
%
explicitly added,
as in case of
ViLBERT \citep{Lu2019VILBERT}, Unicoder-VL \citep{Li2019UNICODERVL}, and ERNIE-ViL \citep{Yu2020ERNIEVIL}.
UNITER \citep{chen2020UNITER} uses -- 
%
in addition to the objects' corners --
the absolute
object
area and the image width and height. OSCAR  \citep{li2020OSCAR}  uses bounding box and image height and width. Only CLIP \citep{radford2021CLIP} does not use PI, although they use another pre-training concept.
See Table~\ref{tab:pi-in-vl} for an overview. To our knowledge, there is no structured analysis of PI in VL models.

\begin{table}[t]
\centering
\begin{tabular}{ll}
\hline
\textbf{PI Type}      & \textbf{Models}\\
      \hline
$\emptyset$          & CLIP           \\
$x1,y1,x2,y2$          & LXMERT, M4C           \\
$x1,y1,x2,y2,\frac{a}{wh}$   & ViLBERT, \\
                    & Unicoder-VL, \\
                    & ERNIE-ViL            \\
$x1,y1,x2,y2,w,h$    & OSCAR            \\
$x1,y1,x2,y2,w,h,a$    & UNITER            \\
\hline
\end{tabular}
\caption{\label{tab:pi-in-vl} Positional information in \mbox{Vision-Language} models. Most models use the upper left and lower right of the object's bounding box ($x1,y1,x2,y2$). Some models add the absolute ($a$) or relative object area ($\frac{a}{wh}$) in combination with the image width ($w$) and height ($h$). The object depth ($d$) is not used and $\emptyset$  denotes no PI.}
\end{table}

Current models use only 2D object information. By introducing depth as a new feature, we represent objects in the 3D space. This
is not only important to be able to define the distances between objects but also to have a more meaningful understanding of the object sizes. Using the area of the bounding box without depth information does not add the actual object size information since the sizes depend on the depth localization of the object.




\section{Evaluation of Positional Information}\label{sec:eval-methods}

To
determine
the capability of current models
with
regard to PI, we
experiment with
three evaluation methods.
Firstly, we perform an intrinsic evaluation
to determine whether
the PI passes through the model.
Secondly, we test if the models
are
capable
of utilizing
PI using the CMM task. 
Lastly, we report extrinsic results for GQA downstream task \citep{Hudson2019GQA} on different data subsets.
We report the results
of the probing experiment 
in Section~\ref{sec:experiments}.

For our experiments, we use four types of PI. An empty set ($\emptyset$) acts as a baseline. Object center values ($x,y$) act as a coarse identification
of
where the object is located.
Moreover, we evaluate $x1,y1,x2,y2$, which is the standard representation of bounding boxes and
is
also often used in VL models.
This PI description contains information about object width, height, and area.
Therefore, we ignore further settings
that add
these types to the input in our evaluation.
Since we are also interested in analyzing depth, we investigate the setting $x1,y1,x2,y2,d$ as well.



\paragraph{Mutual Position Evaluation.} 
In the intrinsic evaluation task, we test if PI is forwarded through the whole model. 
%
We use nine different pairwise classifiers for different mutual positions, which are applied to all detected objects.
LXMERT uses a fixed number of 36 objects as its input. This leads to a total number of $9\times36\times36=11,664$ classifications for each input image.

We use six classifiers
for 2D spatial
relations
(operate on X and Y coordinates) and three
for
depth information (Z coordinate). The tasks are (1) whether the center of one object is more to the left than that of another object, (2) the same if the center is closer to the bottom, (3) whether one object is completely left of the other object (without an overlap), (4) and the same for being completely below the other object, (5) whether one object is completely inside the other bounding box, (6) and if there is no overlap in the X and Y dimension. Regarding depth, the model needs to correctly classify (7) if one object is more in the foreground regarding the median value, (8) if one object is in between the inner 50\% of the other object using all pixel values, and (9) if all depth values of one object are significantly smaller than the values of the other object at a significance level of 0.05 using a $t$-test.

These classification tasks have the same inputs as the \textit{Masked Object Prediction} task (see Section~\ref{sec:base-model}) and are also constructed in the same manner (see Appendix~\ref{app:pi-head}). An overview of the visual pre-training tasks is provided in Figure~\ref{fig:lxmert-pretraining}.
Because this is a probing task, the classification head (PI head) is not updated during pre-training. After the training process has finished, all model parameters are frozen and only the weights in the PI heads are updated for 1 epoch. The average accuracy of all 11,664 classification tasks is reported on the MS COCO validation dataset. In doing so, we evaluate the unimodal capabilities of the model to forward information through the whole Transformer.
The detailed results are presented in Appendix~\ref{app:MPEdetails}.
\jl{Say that detailed results are presented in the Appendix.\pjr{done}}

\jl{11k binary classifiers makes an impression that there are different weights for each task and object pair. There are nine classifiers, but they are used 11k-times.
\pjr{There are 9 classif-tasks, there are 36*36=1296 mutual objects relations, hence 9*1296=11664 classifs} \pjr{13Dec: I rewrote most of it, I think it should be clear now}}
\pjr{Currently I only have results from models with PI pre-training (see Table~\ref{tab:mpe} with Wgt=0). Should I do a fine-tuning for the model w/o PI pre-training? \jl{Sure, this is absolutely necessary to show the pre-training makes a difference at least intrinsically. \pjr{Done in Table~\ref{tab:mpe}}}
}

\paragraph{Contrastive Evaluation on PI using CMM.}  
The CMM classifier can successfully match images and captions (91\% accuracy on the balanced pre-training validation data).
However, this
says little about
the type of information
considered
during the classification.
To
better
assess if PI is
used
by the model, we build a challenge set consisting of pairs of contrastive examples.
We filter the validation data for samples with keywords indicating spatial relation between objects and only keep those which are replaceable by antonyms (see Appendix~\ref{app:oppsite-words}).

We run two evaluation setups:
(1) We replace all image descriptions with a random caption of a different image (following the LXMERT pre-training strategy).
(2) We take the image and for all captions we replace the PI keyword with its antonym, e.g., substitute \textit{background} with \textit{foreground} and vice versa. See Figure~\ref{fig:constrastive-example} for an example. This task determines if the model is able to understand PI in a multimodal fashion.
In both cases, we only have samples with ``no match'' ground truth values (which is our positive class)\footnote{Hence, we have $FP=TN=0$.}, and consequently we report recall only.

\jl{Aren't we interested in precision as well? \pjr{Precision is always 1 by definition, due to FP=TN=0}}


%


\begin{figure}
\noindent\begin{minipage}{0.5\linewidth}
\includegraphics[width=\linewidth]{imgs/COCO_val2014_000000147415.jpg}
\end{minipage}%
\hfill%
\begin{minipage}{0.5\linewidth}\raggedleft
  Original caption: ``A student works on an academic paper at her desk, computer screen glowing in the \underline{background}.''\\
\end{minipage}
\caption{Pre-training data with image and description with a PI keyword. For contrastive evaluation the keyword is replaced by its counterpart (i.e. ``foreground'').}
\label{fig:constrastive-example}
\end{figure}


\paragraph{Downstream Task Evaluation.}  
Finally, we determine the 
performance of the model
on a downstream task. We use GQA, since it is a carefully balanced image question answering dataset, where PI plays a role.
We report the 1- and 5-best accuracy.
Moreover, we evaluate (top 1) accuracy of data subsets where X, Y, and Z coordinates are important.
We do this by selecting questions where specific PI keywords are present (see Appendix~\ref{app:pi-keywords}).

Since keyword search does not work perfectly (e.g., \textit{Which color is the bag on the \underline{back} of the woman?}), we employ zero-shot text classification
using a BART model\footnote{\url{https://huggingface.co/facebook/bart-large-mnli}} \citep{Lewis2019BART}. 
For zero-shot classification we need a candidate label, which is used as input to determine if both texts (i.e. caption and candidate label) fit together.
We experimented with different labels 
and found that the simple
keyword
``position'' works best for our use case.
%


Downstream evaluation is done on the GQA \textit{testdev} split, which has 12,578 samples, hence, an change of 0.1\% is equivalent with approximately 13 more correctly classified samples. For the subsets where X, Y, Z keywords are present, the dataset size is 2,050, 1,203 and 1,349 respectively.
For the zero-shot subset (indicated with P) the sample size is 1,349.

\section{Model and Data}

\subsection{Model}\label{sec:model}

Our experiments are built upon LXMERT -- a Transformer-based model with two separate encoders for image and text modality and one cross-encoder to join both.
LXMERT was the only model in the top-3 leaderboard in both the VQA v2.0 2019 and GQA 2019 challenge, which is why we use this model as the basis for our work. Details are provided in Section~\ref{sec:base-model}.
A detailed description
of
how the object's depth feature is determined is provided in Section~\ref{sec:depth-information}.

\subsubsection{Base Model}\label{sec:base-model}

LXMERT uses Faster R-CNN with ResNet-101 for the object detection task,
originally introduced by \citet{Anderson2018BUTD}. The object detector is trained on Visual Genome \citep{krishna2017VG} predicting 1,600 objects with 400 different attributes (mostly adjectives).
For LXMERT the model extracts the 36 most confident objects with the region-of-interest features $f_j$, the object class $c_j$, attribute $a_j$ and the positional information $p_j$, where $j$ indicates the object indexes $j=1,\ldots,36$. The feature %
map %
($\mathbb{R}^{36\times 2048}$) and
the
bounding box %
coordinates
($\mathbb{R}^{36\times4}$) are passed to two separate linear models with weight matrix $W$ and bias $b$. The output is further processed by two layer normalizations ($\LN$) and finally both results are averaged:
\[\hat{f}_j=\LN(W_Ff_j+b_F) \qquad \hat{p}_j=\LN(W_Pp_j+b_P) \]
\[v_j=(\hat{f}_j + \hat{p}_j)/2\]
\noindent This leads to a unified embedding $v_j \in \mathbb{R}^{36\times768}$ representing
the content of the objects
and the positions at the same time.
The image data is further processed in a BERT-style encoder.

On the language side, the text input is processed in a BERT-style encoder as well.
Both outputs are merged in a cross-modality encoder (X-Enc) and passed to the output heads, where the losses for each pre-training strategy are calculated.
The LXMERT architecture can be investigated in Figure~\ref{fig:lxmert}.

\begin{figure}
\centering
\begin{tikzpicture}[scale=.75, transform shape]
\tikzstyle{every node} = [rectangle, fill=royblue, text=white]
\node (xd)[fill=gray!30, minimum width=15mm, align=center, text=black] at (2, 2.5) {$d_j$};
\node (xp)[minimum width=15mm, align=center, text=white] at (2, 1.5) {{\boldmath$p_j$}};
\node (xf)[minimum width=15mm, align=center] at (2, 0.5) {{\boldmath$f_j$}}; 


\node (center)[minimum width=13mm, minimum height=13mm, align=center, text=white] at (5, 1.5) {\textbf{LXMERT}};

\node(yp)[fill=gray!30, minimum width=13mm, text=black] at (8.75, 3) {$p_j<p_i, d_j<d_i$};
\node(yf)[minimum width=13mm, text=white] at (8, 2) {{\boldmath$f_j$}};
\node(yc)[minimum width=13mm, text=white] at (8, 1) {{\boldmath$c_j$}};
\node(ya)[minimum width=13mm, text=white] at (8, 0) {{\boldmath$a_j$}};

\draw [->] (xf.mid east) -- (center);
\draw [->] (xp.mid east) -- (center);
\draw [->] (xd.mid east)  -- (center) ;
\draw [->] (center) -- (yf.mid west);
\draw [->] (center) -- (yc.mid west);
\draw [->] (center) -- (ya.mid west);
\draw [->] (center) -- (yp.mid west);

\end{tikzpicture}
\caption{Visual components for the pre-training phase (text components omitted). Input data ($f_j, p_j$) to the visual encoder and training targets ($f_j, c_j, a_j$) for LXMERT's pre-training strategies are indicated in blue. Our additional depth data $d_j$ and PI pre-training labels (PIP) are colored gray.}
\label{fig:lxmert-pretraining}
\end{figure}
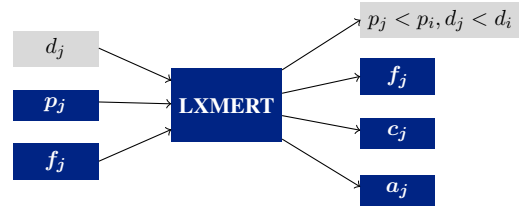

The same pre-training strategies are used, namely \textit{Masked Cross-Modality Language Modeling (MM)}, \textit{Cross-Modality Matching} (CMM), \textit{Visual Question Answering} (VQA), and \textit{Masked Object Prediction}. The last one is composed of three tasks: two classification tasks to predict the object classes and attributes (\textit{ObjClassif}, \textit{AttrClassif}), and a regression task to predict the feature vector (\textit{FeatRegr}). See \citet{Tan2019LXMERT} for all details. Note that all pre-training strategies
explicitly focus
on the object features  $f_j$,  $c_j$, and  $a_j$ and not on the PI. 
The same is true for other VL models listed in Table~\ref{tab:pi-in-vl}. 
See Figure~\ref{fig:lxmert-pretraining} for an illustration of all visual pre-training tasks.

We used the original implementation of LXMERT\footnote{\url{https://github.com/airsplay/lxmert/}} and only made minor changes. 
We introduced dropout with $p=0.1$ in the IQA head.
Further, we tested different training hyperparameters to find a good ratio between model performance and training time.
Our final pre-training model setup has a batch size of 2048 with a learning rate of $10^{-4}$ (with the same learning rate scheduler), the fine-tuning model has a batch size of 32 and a learning rate of $10^{-5}$.
Introducing PyTorch's \textit{DistributedDataParallel} in the code and using 8 
instead of 4 GPUs 
reduced the pre-training time from approximately 8.5 days to 41 hours.
We used the pre-training weights reported in the paper and not in the corresponding repository (see Appendix~\ref{app:pre-training-weights}).


\begin{figure}
\centering
\begin{tikzpicture}[scale=.75, transform shape]
\tikzstyle{every node} = [rectangle, fill=royblue]
\node (img)[minimum width=10mm, draw=black, fill=white] at (0, 1) {Img};
\node (d0)[fill=gray!30, minimum width=15mm, align=center] at (2.45, 1.75) {Depth Estimator}; 
\node (v0)[minimum width=15mm, align=center, text=white] at (2.45, 1) {\textbf{Object Detector}};
\node (v1)[minimum width=13mm, align=center, text=white] at (4.875, 1) {\textbf{VisEnc}};
\node (txt)[minimum width=10mm, draw=black, fill=white] at (0, 0) {Txt};
\node (b)[minimum width=26mm, align=center, text=white] at (4.25, 0) {\textbf{LgnEnc}};
\node (x)[minimum width=13mm, text=white] at (6.75, 0.5) {\textbf{X-Enc}};
\node (tgt)[minimum width=15mm, draw=black, fill=white] at (8.75, 0.5) {Targets};

\draw [->] (img.mid east) -- (v0);
\draw [->] (img.north east) -- (d0.mid west);
\draw [->] (d0.mid east)  -- (v1.north west) ;
\draw [->] (v0) -- (v1);
\draw [->] (v1.mid east) -- (x.north west);
\draw [->] (txt) -- (b);
\draw [->] (b.mid east) -- (x.south west);
\draw [->] (x) -- (tgt) ;

\end{tikzpicture}
\caption{Architecture of the LXMERT model (blue) with depth information extension (gray). LXMERT uses object detection from \citet{Anderson2018BUTD} and has 5 visual, 9 language and 5 cross-modality (X-Enc) layers.}
\label{fig:lxmert}
\end{figure}
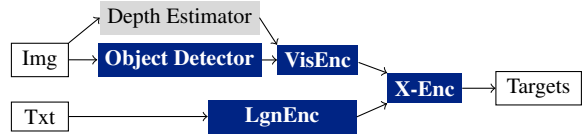

\subsubsection{Depth Information}\label{sec:depth-information}

\begin{figure}
  \centering
  \subfloat[Original image]{\includegraphics[width=0.24\textwidth]{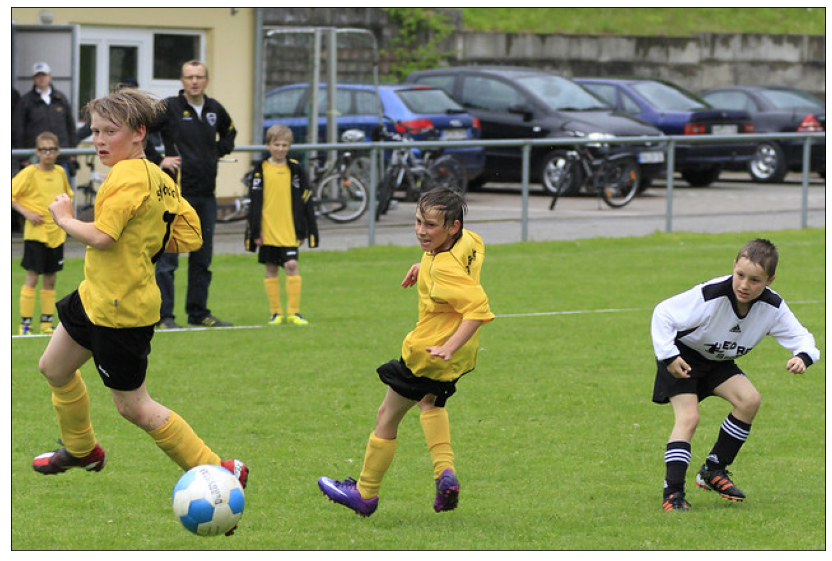}\label{fig:original_cooc}}
  \subfloat[Depth estimation]{\includegraphics[width=0.24\textwidth]{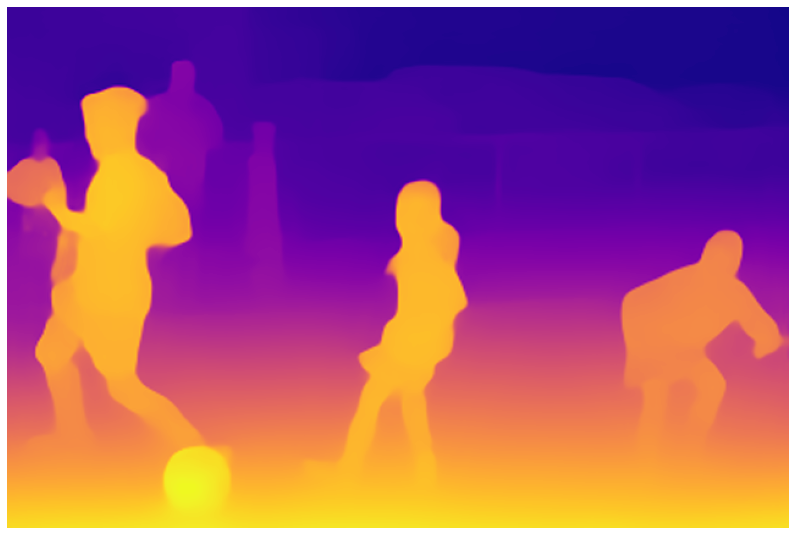}\label{fig:original_midas}}
  \caption{We use a monocular depth estimator to obtain a pixel-level depth prediction. We normalize the output that $0$ (yellow) indicates the value that is at the very front and $1$ for the furthest pixel (violet).}
\end{figure}

The datasets used for training LXMERT do not provide any depth information. To obtain depth values for each pixel in the image, we
used
MiDaS v2.1\footnote{\url{https://pytorch.org/hub/intelisl_midas_v2/}} \citep{Ranftl2019MIDAS} -- a state-of-the-art algorithm for monocular depth estimations. It is trained on diverse datasets from indoor and outdoor environments, containing static and dynamic images and images of different quality. Hence, it
fits
the various picture types in our datasets.
See Figure~\ref{fig:original_cooc} for an original COCO image and Figure~\ref{fig:original_midas} for the depth information provided by the MiDaS model.

The
depth predictions from MiDaS can be any real number.
Large numbers indicate close objects, and small numbers refers to distant objects. We linearly normalized each pixel $x_i$ with $1- \frac{x_i - min(x)}{max(x)}$ to obtain $0$ for the closest pixel and $1$ for the most distant one for each individual image.

Since the rectangular bounding boxes do not surround the objects perfectly, we experimented with the object's center value, the mean and median as heuristic. We finally used the median, due to its robustness. 
%
%
Furthermore, it would be conceivable to additionally take the standard deviation as a measure for uncertainty if the object is on specific depth plane or spans over a larger distance. This issue can be avoided with panoptic segmentation \citep{kirillov2019panoptic}, which we leave to the future work.



\subsection{Data}\label{sec:data}

Following the original LXMERT setup, our models are pre-trained using the MS COCO \citep{Lin2014COCO} and Visual Genome (VG\@; \citealp{krishna2017VG}) data in conjunction with some Visual Question Answering task (VQA). 
There are in total 9.18M image-caption pairs with 180K unique images.
The average sentence length per caption is 10.6 words for MS COCO and 6.2 words for VG. The sentences are short and do not provide
many
details. Using 10 words, only the main occurrence of the image can be described. See examples in Appendix~\ref{app:captions}.


In Table~\ref{tab:keywords-in-datasets}, we show the relative occurrence of PI keywords (see Appendix~\ref{app:pi-keywords}). Pre-training data
do not have
a lot of PI in the captions or questions. Only Y keywords appear more often (11.2\%) in MS COCO and X keywords in VQA (10.7\%). 
This is different in GQA, which we use for downstream evaluation.
In the \textit{train} part, there are many X keywords, but only a few Y and Z keywords. The distribution in the \textit{testdev} set is different. Here, the number of X, Y, and Z questions is high. 
\pjr{Drop "GQA questions are  are automatically generated using a ``Question Engine'' which utilize VG's scene graph which also contains spatial relations." sentence}

LXMERT was also evaluated on VQA v2.0 \citep{balancedvqav2} and NLVR2 \cite{Suhr2019NLVR2}. VQA v2.0 has PI relations (under 3\%),
so
we do not run an analysis on this dataset. NLVR2 has positional relations, but PI keywords are often part of the left/right image assignment and do not indicate objects within an image according to our definition of PI. Moreover, the presence of multiple images rules out a clean analysis of PI.

\begin{table}[t]
\centering
\begin{tabular}{lrrr}
\hline
\textbf{Dataset}  & \textbf{X} & \textbf{Y}     & \textbf{Z}  \\
\hline
MS COCO &  2.9  & 11.2  & 6.5  \\
VG & 3.4  & 3.8  & 4.6  \\
VQA  & 10.7  & 3.3  & 4.0  \\
\hdashline
GQA \textit{train} & 28.4  & 5.3  & 4.9  \\
GQA \textit{testdev} & 16.3   & 9.6  & 10.7  \\
\hline
\end{tabular}
\caption{\label{tab:keywords-in-datasets} Occurrence of positional keywords  in percent in pre-training  (top lines) and downstream datasets (bottom lines).}
\end{table}

\section{Probing Results}\label{sec:experiments}

This section shows the results of
the
experiments described in
Section~\ref{sec:eval-methods}.





\paragraph{Mutual Position Evaluation.}
We determine whether PI can be passed through the model using the classifications of the PI head.
Results are shown in Table~\ref{tab:mpe} (top lines).
The accuracy is
only 80.0\% for no PI, but over 88\% for the remaining types.
This confirms
that the model is able to forward PI through the whole Transformer
layer stack.

Interestingly, the model is often capable
of
correctly
classifying
the mutual position of objects, although PI is not used as
the
model input.
This is most likely due
to
\pjr{adding "most likely", since we cannot proof is? I often use "maybe", "probably" - should I get rid of it? I cannot *proof* that it is really due to correction. }
the high correlation between the object categories and their positions.
For example, ``shoes'' are usually at the bottom and in the foreground.
The object detector is not powerful enough to detect small objects in the background in general.
``Sky'' and ``clouds'' are usually at the top and in the background
of the image.
Detected objects
such as
``kitchen''  or ``office'' often span the whole image width and therefore have their center in the middle of the X axis.
The latent image representation $f_j$ can be used as a  proxy for object types. \pjr{Replaced "is used" with "can be used".}

\pjr{Descriptive analysis of "where does objects occur" is not include now. Do we need this? \jl{Perhaps in the Appendix if you have time.}  \pjr{TODO not sure if I can make it.}}

In addition to that,
we can see that with more PI the accuracy of this task increases by
more than
eight percent points and has a peak at 89.7\% for the input setting $x1,y1,x2,y2,d$.
%
Switching from object centers to bounding boxes only has a minor impact.
Yet, adding depth  improves accuracy on the three Z related tasks (see Appendix~\ref{app:MPEdetails}), which boost the overall performance.

\begin{table}[t]
\centering
\begin{tabular}{clrrr}
\hline
& \textbf{PI }  & \textbf{XYZ} & \textbf{XY} & \textbf{Z}  \\
& \textbf{Input}  & \textbf{Acc} &  \textbf{Acc}  &   \textbf{Acc} \\
      \hline
      \multirow{4}{*}{\rotatebox{90}{Probing}}
& $\emptyset$ &  80.0  & 81.5                      & 77.1 \\
& $x,y$ & 88.5 & 92.1 & 81.1  \\
& $x1,y1,x2,y2$ & 88.7 & 92.4 & 81.3 \\
& $x1,y1,x2,y2,d$ & 89.7  & 92.2 & 84.7 \\
\hline
\color{black}
      \multirow{4}{*}{\rotatebox{90}{Pre-training}}
& $\emptyset$ &  88.2  & 88.9 & 82.1  \\
& $x,y$ & 91.6  &  94.4 & 86.0 \\
& $x1,y1,x2,y2$ & 92.1 &  94.9 & 86.5 \\
& $x1,y1,x2,y2,d$ & 93.9  & 94.8 & 92.2 \\

\hline
\end{tabular}
    \caption{\label{tab:mpe} Mutual Position Classification Evaluation: Mean accuracy of all 9 mutual classification tasks (XYZ), 6 XY tasks, and 3 Z tasks 
    for pre-trained models for different PI inputs. Upper lines for plain LXMERT and lower lines with our version (PIP, CL; see Section~\ref{sec:Improvements}). \pjr{Table\ref{tab:mpe}:  More columns for Mean Acc of XY-Tasks (1-6) and Z-Tasks (7-9)? \jl{It would be nice.}} \pjr{done 7Jan} 
    }
\end{table}


\paragraph{Contrastive Evaluation on PI using CMM.}
To further evaluate the use of PI in VL models, we test if the model can utilize the information using the CMM task.
Table~\ref{tab:permuted-crossmatching} (top lines) shows that the original setting with dissimilar image-text pairs can be predicted almost perfectly -- the recall is always above 96\%.
Hence, this pre-training
strategy
behaves as expected for the normal data provided.
Yet,
the model cannot apply fine-grained details from textual PI.
It
is not capable
of correctly rejecting
that, for example, \textit{``A student works on an academic paper at her desk, computer screen glowing in the \underline{foreground}.''} does not fit to the image from Figure~\ref{fig:constrastive-example}. The recall is steadily below 2\%.

The model is able to pass through PI in the visual Transformer part but is not able to use it in a cross-modal fashion for solving problems.
This is probably due to the fact that fine-grained matching does not play a role during pre-training.
CMM is not constructed as indicated above (i.e., \textit{background} vs. \textit{foreground}) but to select completely dissimilar statements like \textit{"A man sits before a light meal served on the table of a travel trailer''} to the image in Figure~\ref{fig:constrastive-example}.
To overcome this problem, we need more advanced negative sampling, i.e.  captions that are closer to the original image-text pairs.
\jl{more difficult what? \pjr{fine like that?}}

\jl{this face-recognition examples is not necessary. \pjr{removed}}

\begin{table}[t]
\centering
\begin{tabular}{clrr}
\hline
& \textbf{PI}  & \textbf{Permuted}   & \textbf{Permuted} \\[-.2ex]
& \textbf{Input}  & \textbf{caption}   & \textbf{PI words} \\
      \hline
\multirow{4}{*}{\rotatebox{90}{Probing}}
& $\emptyset$  & 97.4 & 1.4 \\
& $x,y$ & 96.5  & 0.3 \\
& $x1,y1,x2,y2$ & 96.8 & 1.7 \\
& $x1,y1,x2,y2,d$ & 97.1 & 1.2\\
      \hline
\color{black}
\multirow{4}{*}{\rotatebox{90}{Pre-training}}
& $\emptyset$  & 96.8  & 78.1 \\
& $x,y$ & 97.7 & 79.5 \\
& $x1,y1,x2,y2$ & 97.7 & 79.3 \\
& $x1,y1,x2,y2,d$ & 97.1 & 79.5\\
\hline
\end{tabular}
    \caption{\label{tab:permuted-crossmatching}Contrastive Evaluation: Recall of the original CMM tasks with random captions (left) and text-image pairs with substituted PI antonyms (right). Upper lines for plain LXMERT and lower lines with our version (PIP, CL; see Section~\ref{sec:Improvements}).}
\end{table}

\paragraph{Downstream Task Evaluation.}
%
We evaluate downstream performance on GQA \textit{testdev} with four different subsets targeting X, Y, Z keywords and general positional (P) samples.
The results (in Table~\ref{tab:gqa-both} top lines) reveal that using any type of PI is better or equally good than not using it (except for Y in $x1, y1, x2, y2$).
Although the improvements are small, they indicate that PI is indeed helpful in this downstream task.

\begin{figure}[t]
  \centering
  \includegraphics[width=.475\textwidth]{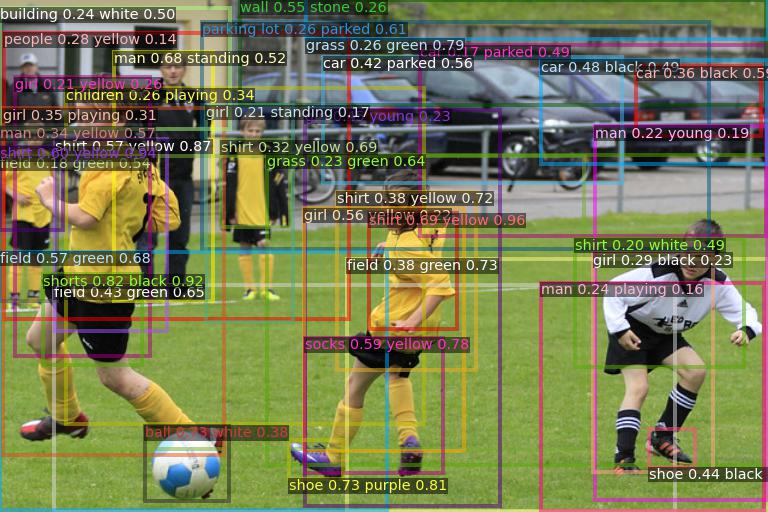}
  \caption{  \label{fig:butd} Bounding box predictions for the 36 objects used in LXMERT. Descriptions contain predicted label and attribute with confidence scores.}
\end{figure}

The best top 1 and X subset results are achieved by $x, y$ input type.
This might \pjr{Same again: Should i add this \textit{uncertainty} word "might" or not? \jl{yes}} be due to the fact that most object relations are distinct and center values are sufficient to track this relationship.
For example, the question ``\textit{Is the boy in white left or right of the ball?}'' is more common than asking ambiguous questions, for example, where bounding boxes intersect (``\textit{Is the left boy in yellow left or right of the ball?}'', see Figure~\ref{fig:butd}).

The PI input $x1, y1, x2, y2,d$ received the best results for the Y and Z subsets.
Although
the
improvements are small, it shows that our new depth feature can help
solve
the Z task.
But also, the improvement on Y can be attributed to the depth input.
Due to the graphical perspective, objects at the top correlate with the background and objects at the bottom with the foreground (see Figure~\ref{fig:original_midas}).
Here, object depth can act as a top/down proxy.

For the downstream evaluation, we need to keep in mind that the underlining object detector is not perfect.
Therefore, we face the issue that objects asked for in the questions are not
always a
part of LXMERT's visual input.
Moreover, our contrastive evaluation scheme shows that LXMERT has difficulties to properly matching image and text representation in a multimodal fashion.
This can explain the small margin of improvements.
The increase of top-1 accuracy
is
not reflected in the
top-5 accuracy.

%
%

\begin{table*}[h]
\centering
\begin{tabular}{clrrrrrr}
\hline
& \textbf{PI Input}   & \textbf{Top 1}  & \textbf{X} & \textbf{Y} & \textbf{Z} & \textbf{P} & \textbf{Top 5} \\
      \hline
\multirow{4}{*}{\rotatebox{90}{Probing}} & $\emptyset$  & 58.1 &  65.7 &      62.0 &      46.4 &      58.0 &      85.0 \\
& $x,y$    &  \textbf{\underline{59.4}} & \underline{\textbf{69.6}} &   62.0 &      49.6 &      \underline{\textbf{60.2}} &   85.0 \\

& $x1,y1,x2,y2$    & 59.0 &   66.2 &      61.8 &      49.4 &      58.9 &      \textbf{85.3} \\
& $x1,y1,x2,y2,d$ & 58.6 &    66.0 &      \underline{\textbf{62.4}} & \underline{\textbf{50.0}}&      58.4 &      85.1 \\
      \hline
\multirow{4}{*}{\rotatebox{90}{Pre-training}} & $\emptyset$     & \textbf{58.8} & \textbf{68.7} &  60.4 & 48.5 & \textbf{59.0} & 85.1  \\
& $x1,y1$         & \textbf{58.8} & \textbf{68.7} &     60.4 &      48.5 &      \textbf{59.0} &   85.1 \\
& $x1,y1,x2,y2,$  & 58.7 & 67.6 &     61.5 &      48.3 &      58.6 &      85.4 \\
& $x1,y1,x2,y2,d$ & 58.7 & 67.8 &     \textbf{62.0} &   \textbf{49.1} &   \textbf{59.0} &   \underline{\textbf{85.8}} \\
\hline

\end{tabular}
    \caption{\label{tab:gqa-both} Evaluation on GQA \textit{testdev}: Model comparison of plain LXMERT models (top lines) and our version with PIP and CL pre-training (bottom lines) for different PI Input types. Evaluation on Top 1 and Top 5 accuracy, moreover on subsets focusing on X, Y, and, Z keywords only and questions that focus on position (P) using zero-shot classification. Underlined numbers indicate the best overall model per column and bold numbers indicate the best model per block and column.%
    }
\end{table*}

\section{From Probing to Pre-training }\label{sec:Improvements}

In the previous
section,
we evaluated the
role of
PI in pre-trained LXMERT\@.
%
%
In this section, we use the probing tasks as a part of model pre-training to improve weaknesses that we identified in the previous section.
Alongside the established strategies, we add two tasks to learn mutual positions and fine-grained PI details in captions utilizing the CMM task. These strategies are elaborated below.

\paragraph{Positional Information Pre-training (PIP).}\label{sec:pi-pretraining}
Currently, all pre-training strategies 
rely
on the visual features ($f_j,c_j,a_j$)
rather than 
on the PI.
%
Only in a small fraction of the pre-training captions and questions positional keywords are present, as Table~\ref{tab:keywords-in-datasets} shows.
Hence, we add a new pre-training strategy which exclusively focuses on PI.

We take the PI head used in Mutual Position Evaluation and add it as a new classification task which is updated during pre-training.
We weight PIP by 10, since the initial loss is noticeably lower than the losses of the other strategies.
Until now only visual representation of the object features, labels and attributes was part of pre-training.
Using PIP, we introduce an explicit unimodal connection between the PI input and the PI output, which was not
previously
available
(see Figure~\ref{fig:lxmert-pretraining}).


\paragraph{Contrastive Learning using CMM (CL).}\label{sec:cl-pretraining}
During pre-training in classical CMM in 50\% of all cases the caption is replaced with another random image description.
This is similar to the main pre-training concept of CLIP.
Yet, doing so, the model only learns to distinguish dissimilar text and images.
There are no small differences in the captions that the model needs to be aware.

In line with Contrastive Evaluation on PI using CMM, we make CMM more complex.
In 50\% of all captions with PI keywords the word is replaced by its counterpart, so that it has to learn fine-grained PI differences during pre-training.
Dissimilar to PIP, this pre-training strategy only affects a small portion of the pre-training samples, since PI keywords are rare.
Yet, it operates on both modalities and hence is able to connect both data types.
This idea can also be extended to other attributes (such as color, material, shape using VG's Scene Graph).


\pjr{Are my results too detailed? \jl{No, this is fine. You should make clear that Top 1 and Top 5 means entire dataset, X,Y,Z,P are subsets.}}

\paragraph{Results.}
Using both pre-training strategies, we train new models for all four PI input types.
We assess the models
using
the same three evaluation schemes as the plain LXMERT model before.

Results of Mutual  Position  Evaluation are shown in Table~\ref{tab:mpe} (bottom lines).
We
observe
an increase in accuracy
for all input types.
The largest is for the empty input type with an accuracy of 88.2\%, indicating the high correlation between feature $f_j$ and position $p_j$.
For the other versions improvements are smaller.
In Table~\ref{tab:MPEdetails} in the Appendix, the accuracies for each of the nine classification tasks are displayed.
The largest increase can be seen for the empty input type with up to 23.1 percent points for task (1) of the 9 mutual position classification tasks.
For classifications based on depth, the best improvements are 9.7 percent points for task (7) and 8.0 percent point for task (9) utilizing $x1, y1, x2, y2, d$.
This shows that the presence of depth is useful as expected.

\jl{You should briefly remind what the task numbers mean. \pjr{done}}

In the original LXMERT version, the probe on Contrastive Evaluation on PI using CMM showed that the model is not able to solve this task successfully.
Recall was steadily below 2 percent.
Introducing the CL pre-training strategy increases matching accuracy to over 78 percent, as shown in Table~\ref{tab:permuted-crossmatching} (bottom lines).
In CMM, we are now able to perform matching between visual and textual representations regarding PI.
As a consequence, we successfully force the model to connect both types in a multimodal manner.

\textbf{Downstream Task Evaluation.}
The third evaluation is the downstream task and
results are shown in Table~\ref{tab:gqa-both} (bottom lines). 
In the two former probes,
%
our extended pre-training
helped the model to solve these tasks.
However, interestingly, this is not the case for GQA evaluation.
The best results for
the
top 1 and subset tasks are obtained by plain LXMERT\@.
Only in the (not official) best 5 accuracy, evaluation our version achieves better results.
One reason for this may be that our PIP weight is too high and needs to be tuned in further studies.
\pjr{Should we talk out problematic PIP weight? Maybe not... \jl{Not necessary.}}

We found that PI has much
less
impact on downstream results as previously thought.
Simple object centers are often sufficient.
Bounding box data, which add object width, height and area, do not add the desired information that the models can utilize.
Adding depth is marginally useful on the Z task, which
suggests
that this feature is useful.

\section{Conclusions}\label{sec:conclusion}

Current VL models make use of different PI inputs without evaluating
their
impact.
In our work, we inspect the effect of such PI input types and 
investigate depth as a new input extension.
In the original setting, the model is able to forward the positional information through the whole Transformer
layer stack,
but it cannot utilize it in the contrastive evaluation and only marginally in the downstream task.
Overall, having any type of PI is helpful, though
object-center values are often sufficient.
However, object features $f_j$ are already good proxies
for
where objects are located.
Because this can be based on spurious correlations, we propose pre-training methods that should make the model rely on PI directly.
%

We introduced two new pre-training strategies.
Firstly, Positional Information Pre-training to ensure that data is passed through the model properly and does not need to rely on feature correlations.
This operates on visual component only and increases performance on the corresponding intrinsic evaluation task.
Moreover, we introduce Contrastive Learning on PI using CMM.
In doing so, we connect PI in the textual and visual modality.
As a result, the model is now able to succeed in the contrastive evaluation task.
%
However, these improvements do not affect the downstream performance on GQA.
%

It is not enough to add different features unchecked, trusting they are properly utilized by the Transformer. In line with BERTology \citep{rogers2020primer,clark2019does,tenney2019bert}, studies are important to understand better what a model is capable.
The same is true for pre-training strategies.
It is not sufficient
to add
new pre-training strategies, although they look promising.
With our probing experiments, we tried to receive a better understanding of the inner workings of LXMERT. 
We see the importance to investigate differences between general concepts and impact on a downstream task.  
\pjr{-> Now positive reference to our Impact...}

We see two major issues for PI in VL models.
Firstly, the pre-training data contains too little fraction of sentences with PI content. Hence, especially the CL pre-training strategy has not enough samples to learn from.
Secondly, the used object detector is not very powerful (see predictions in Figure~\ref{fig:butd}).
Newer detection models like VinVL \citep{zhang2021vinvl} might help to have
a better
image representation, which consequently leverages performance regarding PI context.

\section*{Acknowledgements}

The authors gratefully acknowledge the computing time granted by the Institute for Distributed Intelligent Systems and provided on the GPU cluster Monacum One at the Bundeswehr University Munich.
The work at CUNI was funded by the Czech Science Foundation, grant no.
19-26934X.


\bibliography{anthology,custom,library}

\begin{thebibliography}{29}
\expandafter\ifx\csname natexlab\endcsname\relax\def\natexlab#1{#1}\fi

\bibitem[{Anderson et~al.(2018)Anderson, He, Buehler, Teney, Johnson, Gould,
  and Zhang}]{Anderson2018BUTD}
Peter Anderson, Xiaodong He, Chris Buehler, Damien Teney, Mark Johnson, Stephen
  Gould, and Lei Zhang. 2018.
\newblock Bottom-up and top-down attention for image captioning and visual
  question answering.
\newblock In \emph{Proceedings of the IEEE conference on computer vision and
  pattern recognition}, pages 6077--6086.

\bibitem[{Antol et~al.(2015)Antol, Agrawal, Lu, Mitchell, Batra, Zitnick, and
  Parikh}]{Agrawal2015VQA}
Stanislaw Antol, Aishwarya Agrawal, Jiasen Lu, Margaret Mitchell, Dhruv Batra,
  C~Lawrence Zitnick, and Devi Parikh. 2015.
\newblock {VQA}: Visual question answering.
\newblock In \emph{Proceedings of the IEEE international conference on computer
  vision}, pages 2425--2433.

\bibitem[{Bigham et~al.(2010)Bigham, Jayant, Ji, Little, Miller, Miller,
  Miller, Tatarowicz, White, White et~al.}]{Bigham2010VIZWIZ}
Jeffrey~P Bigham, Chandrika Jayant, Hanjie Ji, Greg Little, Andrew Miller,
  Robert~C Miller, Robin Miller, Aubrey Tatarowicz, Brandyn White, Samual
  White, et~al. 2010.
\newblock Vizwiz: nearly real-time answers to visual questions.
\newblock In \emph{Proceedings of the 23nd annual ACM symposium on User
  interface software and technology}, pages 333--342.

\bibitem[{Chen et~al.(2020)Chen, Li, Yu, El~Kholy, Ahmed, Gan, Cheng, and
  Liu}]{chen2020UNITER}
Yen-Chun Chen, Linjie Li, Licheng Yu, Ahmed El~Kholy, Faisal Ahmed, Zhe Gan,
  Yu~Cheng, and Jingjing Liu. 2020.
\newblock {UNITER}: Universal image-text representation learning.
\newblock In \emph{European conference on computer vision}, pages 104--120.
  Springer.

\bibitem[{Clark et~al.(2019)Clark, Khandelwal, Levy, and
  Manning}]{clark2019does}
Kevin Clark, Urvashi Khandelwal, Omer Levy, and Christopher~D. Manning. 2019.
\newblock \href {https://doi.org/10.18653/v1/W19-4828} {What does {BERT} look
  at? an analysis of {BERT}{'}s attention}.
\newblock In \emph{Proceedings of the 2019 ACL Workshop BlackboxNLP: Analyzing
  and Interpreting Neural Networks for NLP}, pages 276--286, Florence, Italy.
  Association for Computational Linguistics.

\bibitem[{Devlin et~al.(2019)Devlin, Chang, Lee, and Toutanova}]{Devlin2018}
Jacob Devlin, Ming-Wei Chang, Kenton Lee, and Kristina Toutanova. 2019.
\newblock {BERT: Pre-training of Deep Bidirectional Transformers for Language
  Understanding}.
\newblock In \emph{NAACL-HLT}.

\bibitem[{Goyal et~al.(2017)Goyal, Khot, Summers{-}Stay, Batra, and
  Parikh}]{balancedvqav2}
Yash Goyal, Tejas Khot, Douglas Summers{-}Stay, Dhruv Batra, and Devi Parikh.
  2017.
\newblock Making the {V} in {VQA} matter: Elevating the role of image
  understanding in {V}isual {Q}uestion {A}nswering.
\newblock In \emph{Conference on Computer Vision and Pattern Recognition
  (CVPR)}.

\bibitem[{Hu et~al.(2020)Hu, Singh, Darrell, and Rohrbach}]{Hu2019M4C}
Ronghang Hu, Amanpreet Singh, Trevor Darrell, and Marcus Rohrbach. 2020.
\newblock Iterative answer prediction with pointer-augmented multimodal
  transformers for textvqa.
\newblock In \emph{Proceedings of the IEEE/CVF Conference on Computer Vision
  and Pattern Recognition}, pages 9992--10002.

\bibitem[{Hudson and Manning(2019)}]{Hudson2019GQA}
Drew~A Hudson and Christopher~D Manning. 2019.
\newblock {GQA}: A new dataset for real-world visual reasoning and
  compositional question answering.
\newblock In \emph{Proceedings of the IEEE/CVF conference on computer vision
  and pattern recognition}, pages 6700--6709.

\bibitem[{Johnson et~al.(2017)Johnson, Hariharan, van~der Maaten, Fei{-}Fei,
  Zitnick, and Girshick}]{Johnson2016CLEVR}
Justin Johnson, Bharath Hariharan, Laurens van~der Maaten, Li~Fei{-}Fei,
  C.~Lawrence Zitnick, and Ross~B. Girshick. 2017.
\newblock \href {https://doi.org/10.1109/CVPR.2017.215} {{CLEVR:} {A}
  diagnostic dataset for compositional language and elementary visual
  reasoning}.
\newblock In \emph{2017 {IEEE} Conference on Computer Vision and Pattern
  Recognition, {CVPR} 2017, Honolulu, HI, USA, July 21-26, 2017}, pages
  1988--1997. {IEEE} Computer Society.

\bibitem[{Ke et~al.(2021)Ke, He, and Liu}]{Ke2020rethinking}
Guolin Ke, Di~He, and Tie-Yan Liu. 2021.
\newblock Rethinking {P}ositional {E}ncoding in {L}anguage {P}re-training.
\newblock In \emph{International Conference on Learning Representations}.

\bibitem[{Kirillov et~al.(2019)Kirillov, He, Girshick, Rother, and
  Doll{\'a}r}]{kirillov2019panoptic}
Alexander Kirillov, Kaiming He, Ross Girshick, Carsten Rother, and Piotr
  Doll{\'a}r. 2019.
\newblock Panoptic segmentation.
\newblock In \emph{Proceedings of the IEEE/CVF Conference on Computer Vision
  and Pattern Recognition}, pages 9404--9413.

\bibitem[{Krishna et~al.(2017)Krishna, Zhu, Groth, Johnson, Hata, Kravitz,
  Chen, Kalantidis, Li, Shamma et~al.}]{krishna2017VG}
Ranjay Krishna, Yuke Zhu, Oliver Groth, Justin Johnson, Kenji Hata, Joshua
  Kravitz, Stephanie Chen, Yannis Kalantidis, Li-Jia Li, David~A Shamma, et~al.
  2017.
\newblock Visual genome: Connecting language and vision using crowdsourced
  dense image annotations.
\newblock \emph{International Journal of Computer Vision}, 123(1):32--73.

\bibitem[{Lewis et~al.(2020)Lewis, Liu, Goyal, Ghazvininejad, Mohamed, Levy,
  Stoyanov, and Zettlemoyer}]{Lewis2019BART}
Mike Lewis, Yinhan Liu, Naman Goyal, Marjan Ghazvininejad, Abdelrahman Mohamed,
  Omer Levy, Veselin Stoyanov, and Luke Zettlemoyer. 2020.
\newblock \href {https://doi.org/10.18653/v1/2020.acl-main.703} {{BART}:
  Denoising sequence-to-sequence pre-training for natural language generation,
  translation, and comprehension}.
\newblock In \emph{Proceedings of the 58th Annual Meeting of the Association
  for Computational Linguistics}, pages 7871--7880, Online. Association for
  Computational Linguistics.

\bibitem[{Li et~al.(2020{\natexlab{a}})Li, Duan, Fang, Gong, and
  Jiang}]{Li2019UNICODERVL}
Gen Li, Nan Duan, Yuejian Fang, Ming Gong, and Daxin Jiang. 2020{\natexlab{a}}.
\newblock Unicoder-vl: A universal encoder for vision and language by
  cross-modal pre-training.
\newblock In \emph{Proceedings of the AAAI Conference on Artificial
  Intelligence}, volume~34, pages 11336--11344.

\bibitem[{Li et~al.(2020{\natexlab{b}})Li, Yin, Li, Hu, Zhang, Zhang, Wang, Hu,
  Dong, Wei, Choi, and Gao}]{li2020OSCAR}
Xiujun Li, Xi~Yin, Chunyuan Li, Xiaowei Hu, Pengchuan Zhang, Lei Zhang, Lijuan
  Wang, Houdong Hu, Li~Dong, Furu Wei, Yejin Choi, and Jianfeng Gao.
  2020{\natexlab{b}}.
\newblock Oscar: Object-semantics aligned pre-training for vision-language
  tasks.
\newblock \emph{ECCV 2020}.

\bibitem[{Lin et~al.(2014)Lin, Maire, Belongie, Hays, Perona, Ramanan,
  Doll{\'a}r, and Zitnick}]{Lin2014COCO}
Tsung-Yi Lin, Michael Maire, Serge Belongie, James Hays, Pietro Perona, Deva
  Ramanan, Piotr Doll{\'a}r, and C.~Lawrence Zitnick. 2014.
\newblock Microsoft coco: Common objects in context.
\newblock In \emph{Computer Vision -- ECCV 2014}, pages 740--755, Cham.
  Springer International Publishing.

\bibitem[{Lu et~al.(2019)Lu, Batra, Parikh, and Lee}]{Lu2019VILBERT}
Jiasen Lu, Dhruv Batra, Devi Parikh, and Stefan Lee. 2019.
\newblock Vilbert: Pretraining task-agnostic visiolinguistic representations
  for vision-and-language tasks.
\newblock \emph{Advances in neural information processing systems}, 32.

\bibitem[{Radford et~al.(2021)Radford, Kim, Hallacy, Ramesh, Goh, Agarwal,
  Sastry, Askell, Mishkin, Clark, Krueger, and Sutskever}]{radford2021CLIP}
Alec Radford, Jong~Wook Kim, Chris Hallacy, Aditya Ramesh, Gabriel Goh,
  Sandhini Agarwal, Girish Sastry, Amanda Askell, Pamela Mishkin, Jack Clark,
  Gretchen Krueger, and Ilya Sutskever. 2021.
\newblock \href {http://arxiv.org/abs/2103.00020} {Learning transferable visual
  models from natural language supervision}.

\bibitem[{Ranftl et~al.(2020)Ranftl, Lasinger, Hafner, Schindler, and
  Koltun}]{Ranftl2019MIDAS}
Ren\'{e} Ranftl, Katrin Lasinger, David Hafner, Konrad Schindler, and Vladlen
  Koltun. 2020.
\newblock Towards robust monocular depth estimation: Mixing datasets for
  zero-shot cross-dataset transfer.
\newblock \emph{IEEE Transactions on Pattern Analysis and Machine Intelligence
  (TPAMI)}.

\bibitem[{Rogers et~al.(2020)Rogers, Kovaleva, and
  Rumshisky}]{rogers2020primer}
Anna Rogers, Olga Kovaleva, and Anna Rumshisky. 2020.
\newblock \href {https://doi.org/10.1162/tacl_a_00349} {A primer in
  {BERT}ology: What we know about how {BERT} works}.
\newblock \emph{Transactions of the Association for Computational Linguistics},
  8:842--866.

\bibitem[{Shaw et~al.(2018)Shaw, Uszkoreit, and Vaswani}]{shaw-etal-2018-self}
Peter Shaw, Jakob Uszkoreit, and Ashish Vaswani. 2018.
\newblock \href {https://doi.org/10.18653/v1/N18-2074} {Self-attention with
  relative position representations}.
\newblock In \emph{Proceedings of the 2018 Conference of the North {A}merican
  Chapter of the Association for Computational Linguistics: Human Language
  Technologies, Volume 2 (Short Papers)}, pages 464--468, New Orleans,
  Louisiana. Association for Computational Linguistics.

\bibitem[{Suhr et~al.(2019)Suhr, Zhou, Zhang, Zhang, Bai, and
  Artzi}]{Suhr2019NLVR2}
Alane Suhr, Stephanie Zhou, Ally Zhang, Iris Zhang, Huajun Bai, and Yoav Artzi.
  2019.
\newblock \href {https://doi.org/10.18653/v1/P19-1644} {A corpus for reasoning
  about natural language grounded in photographs}.
\newblock In \emph{Proceedings of the 57th Annual Meeting of the Association
  for Computational Linguistics}, pages 6418--6428, Florence, Italy.
  Association for Computational Linguistics.

\bibitem[{Tan and Bansal(2019)}]{Tan2019LXMERT}
Hao Tan and Mohit Bansal. 2019.
\newblock \href {https://doi.org/10.18653/v1/D19-1514} {{LXMERT}: Learning
  cross-modality encoder representations from transformers}.
\newblock In \emph{Proceedings of the 2019 Conference on Empirical Methods in
  Natural Language Processing and the 9th International Joint Conference on
  Natural Language Processing (EMNLP-IJCNLP)}, pages 5100--5111, Hong Kong,
  China. Association for Computational Linguistics.

\bibitem[{Tenney et~al.(2019)Tenney, Das, and Pavlick}]{tenney2019bert}
Ian Tenney, Dipanjan Das, and Ellie Pavlick. 2019.
\newblock \href {https://doi.org/10.18653/v1/P19-1452} {{BERT} {R}ediscovers
  the {C}lassical {NLP} pipeline}.
\newblock In \emph{Proceedings of the 57th Annual Meeting of the Association
  for Computational Linguistics}, pages 4593--4601, Florence, Italy.
  Association for Computational Linguistics.

\bibitem[{Vaswani et~al.(2017)Vaswani, Shazeer, Parmar, Uszkoreit, Jones,
  Gomez, Kaiser, and Polosukhin}]{Vaswani2017}
Ashish Vaswani, Noam Shazeer, Niki Parmar, Jakob Uszkoreit, Llion Jones,
  Aidan~N Gomez, {\L}ukasz Kaiser, and Illia Polosukhin. 2017.
\newblock Attention is all you need.
\newblock In \emph{Advances in neural information processing systems}, pages
  5998--6008.

\bibitem[{Wang and Chen(2020)}]{Wang2020what}
Yu-An Wang and Yun-Nung Chen. 2020.
\newblock \href {https://doi.org/10.18653/v1/2020.emnlp-main.555} {What do
  position embeddings learn? an empirical study of pre-trained language model
  positional encoding}.
\newblock In \emph{Proceedings of the 2020 Conference on Empirical Methods in
  Natural Language Processing (EMNLP)}, pages 6840--6849, Online. Association
  for Computational Linguistics.

\bibitem[{Yu et~al.(2021)Yu, Tang, Yin, Sun, Tian, Wu, and
  Wang}]{Yu2020ERNIEVIL}
Fei Yu, Jiji Tang, Weichong Yin, Yu~Sun, Hao Tian, Hua Wu, and Haifeng Wang.
  2021.
\newblock Ernie-vil: Knowledge enhanced vision-language representations through
  scene graphs.
\newblock In \emph{Proceedings of the AAAI Conference on Artificial
  Intelligence}, volume~35, pages 3208--3216.

\bibitem[{Zhang et~al.(2021)Zhang, Li, Hu, Yang, Zhang, Wang, Choi, and
  Gao}]{zhang2021vinvl}
Pengchuan Zhang, Xiujun Li, Xiaowei Hu, Jianwei Yang, Lei Zhang, Lijuan Wang,
  Yejin Choi, and Jianfeng Gao. 2021.
\newblock Vinvl: Making visual representations matter in vision-language
  models.
\newblock \emph{CVPR 2021}.

\end{thebibliography}
\bibliographystyle{acl_natbib}

\appendix
\section{Appendix}
\pjr{Only "A" Appendix or several letters? \jl{It does not matter.}}

\subsection{PI Classification Head}
\label{app:pi-head}
The PI head is build up in the same manner as the other visual heads, i.e.
\texttt{Dense $\rightarrow$ Activation $\rightarrow$ \mbox{Layer Normalization} $\rightarrow$ Dropout $\rightarrow$ Dense}.
\pjr{Not really nice, but I don't what might be better\pjr{I thinks it's fine so}}


\subsection{PI Antonyms}
\label{app:oppsite-words}

For Contrastive Evaluation, we replace some PI keywords with its antonyms.

We substitute \textit{left} with \textit{right}, \textit{above} with \textit{below}, \textit{under} with \textit{over}, \textit{foreground} with \textit{background}, \textit{before} with \textit{behind} and vice versa.


\subsection{PI Keywords}\label{app:pi-keywords}

In Table~\ref{tab:xyz-keywords} we list all PI keywords used in our evaluations.

\begin{table}[h]
\centering
\begin{tabular}{ll}
\hline
\textbf{Dim.}      & \textbf{Keywords}  \\
      \hline
X    &  left, right, beside, besides, alongside, side \\
Y  &  top, down, above, below, under, beneath, \\
   & underneath, over, beyond, overhead \\
Z  & behind, front, rear, back, ahead, before, \\
 & foreground, background, before, forepart, \\
   & far end, hindquarters\\
\hline
\end{tabular}
\caption{\label{tab:xyz-keywords} Overview of positional keywords regarding dimension.}
\end{table}

\subsection{Pre-training Weights}
\label{app:pre-training-weights}

In Table~\ref{tab:weights} we compare pre-training weights from LXMERT paper \citep{Tan2019LXMERT} and the repository version (\url{https://github.com/airsplay/lxmert/}). 

\begin{table}[h]
\centering
\begin{tabular}{lrrrrrr}
\hline
\textbf{Version}      & \rot{\textbf{MLM}} & \rot{\textbf{CMM}} & \rot{\textbf{ObjClassif}} & \rot{\textbf{AttrClassif}{ }} & \rot{\textbf{FeatRegr}} & \rot{\textbf{VQA}}   \\
      \hline
Paper   & 1  & 1 & 1 & 1 & 1 & 1 \\
Repository  &  1  & 1 & $6.\overline{6}$ & $6.\overline{6}$ & $6.\overline{6}$ & 1 \\
\hline
\end{tabular}
\caption{\label{tab:weights} Overview of pre-training weights in publication and GitHub version. \pjr{Tan2019 is just a github repo. Cite it like that? Or with "Paper version", "Code version"?}}
\end{table}

\subsection{Text Examples}
\label{app:captions}

In Table~\ref{tab:captions} we provide examples from pre-training and downstream tasks with highlighted keywords.

\begin{table*}
\centering
\begin{tabular}{llr}
\hline
\textbf{Dataset}      & \textbf{Example}   & \textbf{Length} \\
\hline
MS COCO &  A very clean and well decorated empty bathroom & 8 \\
        &  A panoramic view of a kitchen and all of its appliances. & 11\\
        &  Surfers waiting for the \textit{right} wave to ride. &  8 \\
        & Two dogs are laying \textit{down} \textbf{next} to each other.& 9 \\
        & A red stop sign with a Bush bumper sticker \textbf{under} the word stop.& 13 \\
\hline
VG      & separate kitchen areas in a home & 6 \\
        & older red Volkswagen Beetle car & 5\\
        & a woman walking \textit{down} the sidewalk & 6 \\
        & A bag in the woman's \textbf{left} hand & 7 \\
        & stones \textbf{under} wood bench & 4 \\
\hline
GQA     & Are there both a television and a chair in the picture? & 11\\
        & That car is what color? &  5 \\
        & On which \textbf{side} of the picture is the lamp?  & 9 \\
        & Is the table to the \textbf{left} or to the \textbf{right} of the appliance in the center? & 16 \\
        & Is there a bookcase \textbf{behind} the yellow flowers?  & 8\\

\hline
\end{tabular}
\caption{\label{tab:captions} Text examples from different datasets with word counts. Italic stands for PI keywords that are wrongly selected and bold words are correctly detected. \pjr{E.g. "center" is not in my PI keyword list. }}
\end{table*}

\subsection{Mutual Positional Evaluation Details}
\label{app:MPEdetails}

\jl{Could you remind here what the tasks were?}

In Table~\ref{tab:MPEdetails} we provide detailed results for all 9 mutual PI tasks. Tasks (1)-(6) relate to X and Y coordinates and tasks (7)-(9) to Z coordinates.
The numbering is explained in Section~\ref{sec:eval-methods}.

\begin{table*}
\centering
\begin{tabular}{lrrrrrrrrr}
\hline
\textbf{PI Input} &    \textbf{(1)} &    \textbf{(2)} &    \textbf{(3)} &   \textbf{(4)} &    \textbf{(5)} &    \textbf{(6)} &    \textbf{(7)} &    \textbf{(8)} &  \textbf{(9)} \\
\hline
$\emptyset$      & 65.0 & 84.1 & 82.1 & 89.9 & 95.6 & 72.3 & 77.7 & 75.3 & 78.4 \\
$x,y$        & 95.1 & 95.6 & 96.2 & 96.1 & 95.8 & 74.1 & 83.3 & 75.7 & 84.4 \\
$x1,y1,x,2,y2$    & 94.3 & 95.2 & 96.8 & 97.0 & 96.0 & 75.0 & 83.5 & 75.8 & 84.6 \\
$x1,y1,x,2,y2,d$ & 94.0 & 95.0 & 96.6 & 96.8 & 96.0 & 74.9 & 88.7 & 76.3 & 89.1 \\
\hline
$\emptyset$      & 88.1 & 89.4 & 92.6 & 93.5 & 95.9 & 74.1 & 83.9 & 77.7 & 84.8 \\
$x,y$       & 98.7 & 98.8 & 98.3 & 98.3 & 96.1 & 75.9 & 89.3 & 78.4 & 90.4 \\
$x1,y1,x,2,y2$     & 98.8 & 98.9 & 98.7 & 99.5 & 96.3 & 77.0 & 89.7 & 78.9 & 90.9 \\
$x1,y1,x,2,y2,d$ & 98.9 & 98.9 & 98.6 & 99.0 & 96.3 & 77.2 & 98.4 & 81.0 & 97.1 \\
\hline
\end{tabular}
\caption{\label{tab:MPEdetails} Average accuracy per classification task (1-9) in Mutual Positional Evaluation for plain LXMERT (top lines) and our version (bottom lines).}
\end{table*}

\end{document}